\documentclass[11pt]{article}

\usepackage[preprint]{acl}

\usepackage{times}
\usepackage{latexsym}

\usepackage[T1]{fontenc}

\usepackage[utf8]{inputenc}

\usepackage{microtype}

\usepackage{inconsolata}

\usepackage{graphicx}

\usepackage{hyperref}
\usepackage{listings}
\usepackage{xcolor}
\usepackage{tcolorbox}
\usepackage{amsmath}
\usepackage{amsfonts}
\usepackage{nicefrac}
\usepackage{multirow}
\usepackage{svg}
\tcbuselibrary{breakable, skins}
\usepackage{booktabs}
\usepackage{url}
\usepackage{xspace}
\usepackage{float}
\usepackage{textcomp}
\usepackage{amssymb}
\usepackage{makecell}
\usepackage{threeparttable}
\usepackage{wasysym}
\usepackage{pifont}
\usepackage{subfigure}
\usepackage{subcaption}
\usepackage{array}
\usepackage[table]{xcolor}
\usepackage{xparse}
\usepackage{caption}

\newcommand{\mname}{\text{TrajSelector}}  
\definecolor{darkgreen}{rgb}{0, 0.5, 0}

\definecolor{navyblue}{HTML}{0071BC}

\NewDocumentCommand{\passk}{O{green!50!black}m}{%
    \textcolor{#1}{\textbf{#2}}%
}

\title{
\includegraphics[height=\baselineskip]{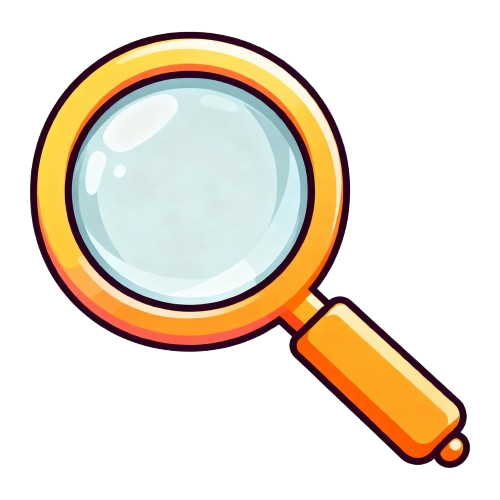}
\emph{\mname{}}: Harnessing Latent Representations \\ for Efficient and Effective Best-of-$N$ in Large Reasoning Model
}

\author{
 \textbf{Bin Yu\textsuperscript{1,2}},
 \textbf{Xinming Wang\textsuperscript{2,4}},
 \textbf{Shijie Lian\textsuperscript{2,5}},
 \textbf{Haotian Li \textsuperscript{1}},
\\
 \textbf{Changti Wu\textsuperscript{2,6}},
 \textbf{Ruina Hu\textsuperscript{1,2}},
 \textbf{Bailing Wang\textsuperscript{1}},
 \textbf{Yuliang Wei\textsuperscript{1,\thanks{Corresponding author}}},
 \textbf{Kai Chen \textsuperscript{2,3,\footnotemark[1]}}
\\
\\
 \textsuperscript{1}Harbin Institute of Technology,
 \textsuperscript{2}Zhongguancun Academy
\\
 \textsuperscript{3}Zhongguancun Institute of Artificial Intelligence
\\
 \textsuperscript{4}Institute of Automation, Chinese Academy of Sciences
\\
 \textsuperscript{5}Huazhong University of Science and Technology,
 \textsuperscript{6}East China Normal University
}

\begin{document}
\maketitle

\begin{abstract}

Large language models (LLMs) have shown remarkable progress in complex reasoning tasks, largely enabled by test-time scaling (TTS) paradigms that allocate additional compute during inference. Among these, external TTS—particularly the Best-of-$N$ selection paradigm—yields scalable performance improvements by selecting from multiple independently generated reasoning trajectories. However, this approach faces key limitations: (i) the high computational overhead of deploying process reward models, (ii) the underutilization of the LLM’s intrinsic latent representations. We introduce \emph{\textbf{\mname{}}}, an efficient and effective Best-of-$N$ framework that exploit the hidden states in the sampler LLM for process-level scoring. A lightweight verifier (with only 0.6B parameters) evaluates the quality of step-wise trajectory, and then aggregates these scores to identify the optimal reasoning trajectory. Our framework employs a fully data-driven, end-to-end training recipe that eliminates reliance on massive step-level annotations. Experiential results across five benchmarks demonstrate that \emph{\mname{}} delivers consistent performance gains. In Best-of-32 settings, it surpasses majority voting by 4.61\% accuracy and outperforms existing process reward models by 4.31\% to 12.21\%, all while maintaining lower inference costs.
Project website: \href{https://zgca-ai4edu.github.io/TrajSelector}{https://zgca-ai4edu.github.io/TrajSelector}.

\end{abstract}

\section{Introduction}
\label{sec:intro}

Large language models (LLMs) have achieved remarkable progress in domains such as mathematical reasoning over the past two years \citep{Gemini2.5,DeepSeekMath,ScientificAgentSurvey}. A key driver of these advancements is the emergence of the test-time scaling paradigm \citep{guo2025deepseek,s1,CognitionEngineering}, which boosts model performance by allocating additional computational resources during the inference phase. Test-time scaling (TTS) can be categorized into two complementary forms: (i) \textbf{Internal TTS} \citep{chain-of-thought,LS-Mixture} achieves this by extending the model's reasoning of longer chain-of-thoughts; (ii) \textbf{External TTS} \citep{Parallel-R1,MetaStone-S1,DeepConf,OpenAI-o3-Breakthrough} does so by having the model exploring multiple reasoning solutions in parallel.

\begin{figure}[!t]
  \centering
  \includegraphics[width=0.48\textwidth]{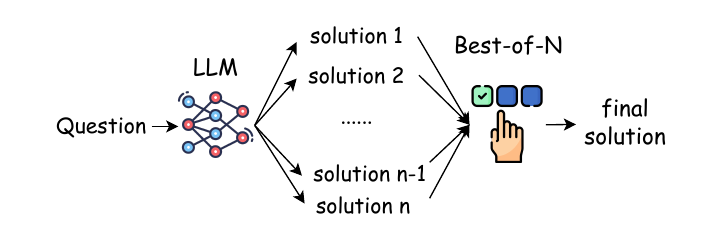}
  \caption{Best-of-$N$ selection method illustration.}
  \label{fig:best-of-n}
\end{figure}

Our work centers on harnessing external TTS to enhance model performance. Although generating multiple independent solutions enables parallelized computation, the key challenge lies in answer aggregation, formally modeled as a \textbf{Best-of-$N$} selection problem (Figure~\ref{fig:best-of-n}). Existing approaches fall into two categories: (i) the use of independent process reward models (PRMs) as external verifiers to select among reasoning trajectories \citep{ReasonEval,ReasonFlux-PRM,EurusPRM}; (ii) the exploitation of intrinsic model states for correctness evaluation \citep{Self-Consistency,TTRL,VL-TTRL,Beyond28}. However, both methods encounter key limitations: standalone PRMs often require computationally expensive deployments at the 7B scale \citep{Qwen2.5-Math-PRM-7B,ReasonFlux-PRM}, while state-based methods suffer from inconsistent performance and reliability issues across diverse tasks \citep{NoFreeLunch,AggLM}.

We identify two primary bottlenecks in existing Best-of-$N$ methods: (i) high-quality verifiers are typically large and computationally intensive, and their internal representations are often misaligned with those of the sampler LLM; (ii) training PRMs requires generally relies on costly step-level annotations. These constraints hinder the practicality of deploying process verifiers in real-world settings. Overcoming these challenges requires a lightweight verifier that effectively exploit the sampler LLM’s intrinsic latent representation, coupled with a training strategy that eliminates the need for step-level supervision.

Motivated by these challenges, we propose {\mname{}}, an efficient and effective Best-of-$N$ framework that explot the latent representations inherent in the sampler LLM for solution selection. Our framework requires only a minimal-parameter LLM to function as a process verifier. The core idea is to repurpose the last hidden states from the sampler—rich in introspective signals—for step-level scoring. By coupling generation and evaluation at the representation level, {\mname{}} facilitates accurate reasoning assessment with a minimal number of additional parameters, eliminating the need for a large standalone verifier.


\begin{figure}[!t]
\centering
\includegraphics[width=0.48\textwidth]{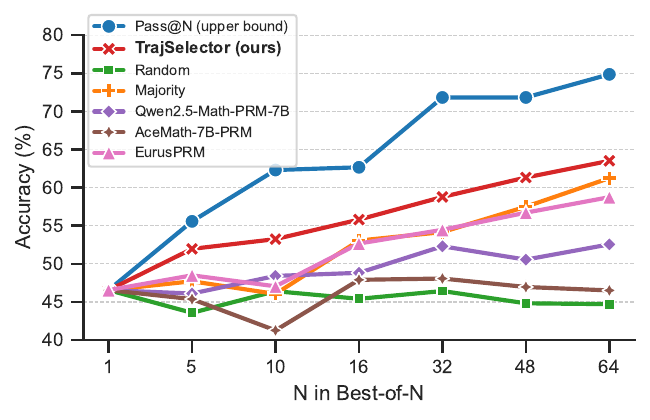}
\captionsetup{skip=3pt}
\caption{Best-of-$N$ Scaling Curve. The y-axis represents the average accuracy of different methods across the 5 benchmarks in the experimental section. \emph{\textbf{\mname{}}} achieves robust performance improvement as $N$ increases.}
\label{fig:best-of-n-scaling}
\end{figure}


We construct the training dataset using OpenR1-Math-220K \citep{openr1} and DeepMath-103K \citep{deepmath} as data sources. To eliminate reliance on process-level annotations, we adopt a data-driven training recipe for \emph{\mname{}}. During training, the sampler LLM is kept frozen, and only the lightweight process verifier is updated. Compared to full-scale PRM training, this approach demands significantly fewer computational resources while achieving superior Best-of-$N$ performance. As illustrated in Figure~\ref{fig:best-of-n-scaling}, {\mname{}} delivers consistent performance improvements across a range of Best-of-$N$ settings ($N \in [1, 64]$). In the Best-of-32 setting, our method surpasses Majority Voting by 4.61\% in accuracy and outperforms other process reward models by 4.31\% to 12.21\%, using only a 0.6B-parameter verifier.

Our main contributions are as follows:
\begin{itemize}
    \item We present \emph{\textbf{\mname{}}}, which reuses the sampler model’s step-final hidden states as self-reflective signals for a lightweight verifier to perform step-wise scoring and pooled trajectory evaluation, enabling efficient Best-of-$N$ selection. This design achieves test-time gains with minimal parameter overhead.
    \item We present an end-to-end, data-driven training paradigm for a process verifier that eliminates the need for step-level label annotations. Combined with a compact model architecture, this approach significantly reduces the training cost of the verifier.
    \item \mname{} demonstrates a favorable accuracy–compute trade-off compared to majority voting and heavy verifiers, while maintaining robust performance across varing Best-of-$N$ settings. This establishes a practical foundation for future work in external TTS optimization.
\end{itemize}

\section{Related Work}
\label{sec:related}

\subsection{Test-Time Scaling (TTS)}

Test-Time Scaling improves model reasoning by allocating additional compute at inference, and has recently become central with the success of OpenAI o1 \citep{jaech2024openai}.
Internal TTS scales the depth of reasoning within the model via extended chain-of-thought (CoT).
Systems like OpenAI o1 \citep{jaech2024openai} and DeepSeek-R1 \citep{guo2025deepseek} explicitly integrate structured reasoning traces, enabling decomposition and self-correction \citep{jin2024impact,yeo2025demystifying}.
Complementarily, external TTS scales the breadth of inference by generating and evaluating multiple reasoning trajectories. The most prevalent paradigm among these involves parallel sampling from the model, followed by Best-of-$N$ aggregation of solutions \citep{Self-Consistency,ReasonWithoutCoT,MetaStone-S1,Parallel-R1}. \citet{Self-Consistency} and \citet{TTRL} employ a majority voting scheme to select the final solution from candidates. \citep{ReasonWithoutCoT} demonstrates that multiple sampling without explicit reasoning outperforms a single reasoning process augmented with long chain-of-thought. \citet{MetaStone-S1} employs two linear layers for process scoring; however, it requires reinforcement learning to dynamically adjust the parameters of the sampler LLM during training, which may induce catastrophic forgetting in the policy model. In contrast, our method does not require parameter modifications to the sampler LLM, thereby avoiding training failures arising from issues in the quality and distribution of post-training data. 

\subsection{Process Reward Model (PRM)}

For selecting the most likely correct answer from multiple candidates, the mainstream approach employs PRMs for scoring and selection. Representative open efforts include Qwen2.5-Math-PRM-7B \citep{Qwen2.5-Math-PRM-7B}, which reports step-level gains over strong baselines at a comparable scale, and the PRM800K-tuned PRM Qwen2.5-Math-7B-PRM800K \citep{ProcessBench}, both emphasizing the value of large, process-labeled corpora for math reasoning.
To reduce human labeling, works such as Math-Shepherd \citep{Math-Shepherd} and ReasonEval-7B \citep{ReasonEval} automate stepwise assessment by external tools. 
Beyond stepwise scoring, ReasonFlux-PRM-7B \citep{ReasonFlux-PRM} introduces trajectory-aware supervision for improving Best-of-$N$ TTS.
AceMath-7B-RM \citep{AceMath} provides strong math reward models and establish a practical baseline. 
Further, EurusPRM \citep{EurusPRM} advances implicit and online process rewards aiming to scalable PRMs for external TTS.
However, deploying the aforementioned PRMs for Best-of-$N$ selection necessitates the independent deployment of a large model (approximately 7B parameters), whose scale approaches that of the sampler LLM, thereby incurring substantial increases in deployment and inference costs. In stark contrast, our method requires only an additional 0.6B tiny LLM to serve as the process verifier.

\section{Problem Statement}
We consider the task of reasoning-aware answer generation using LLMs. Given a natural language query $x$, the model $\mathcal{M}$ is required to generate a final answer $\hat{r}$ accompanied by a sequence of $T$ intermediate reasoning steps $\tau = (s_1, s_2, \ldots, s_T)$ that reflect the model’s internal decision-making trajectory:
\begin{equation}
\tau \sim \mathcal{M}(\cdot | x), ~~ \hat{r} \sim \mathcal{M}(\cdot | \tau, x)
\end{equation}

Each reasoning step $s_t$ corresponds to a discrete cognitive operation—such as deduction, retrieval, transformation, or hypothesis refinement—that cumulatively constructs the path from the query to the final answer.

The external test-time scaling (TTS) problem we address involves improving LLM performance post-training by leveraging multiple independently generated responses at inference time. Each data point is represented as a tuple $(x, r)$, consisting of a query and its corresponding ground-truth answer. To assess the correctness of a generated response $\hat{r}$, we define a binary label:
\begin{equation}
y = \mathbb{I}[r = g(\hat{r})]
\end{equation}
where $\mathbb{I}[\cdot]$ is the indicator function, returning 1 if the model’s response exactly matches the ground truth $r$, and 0 otherwise. $g(\cdot)$ is used to extract the answer to the query from the response $\hat r$, often implemented in a rule-based manner.

The Best-of-$N$ strategy serves as a foundational technique in external TTS. Rather than relying on a single output, the model generates $N$ independent responses $\hat{r}_1, \hat{r}_2, \ldots, \hat{r}_N$, each evaluated for correctness with labels $y_1, \ldots, y_N$. Under this paradigm, the model $\mathcal{M}$ is considered to have answered the query $x$ correctly if at least one $y_n = 1$.
\begin{figure*}[!htbp]
    \centering
    \includegraphics[trim=0 15pt 0 10pt, clip]{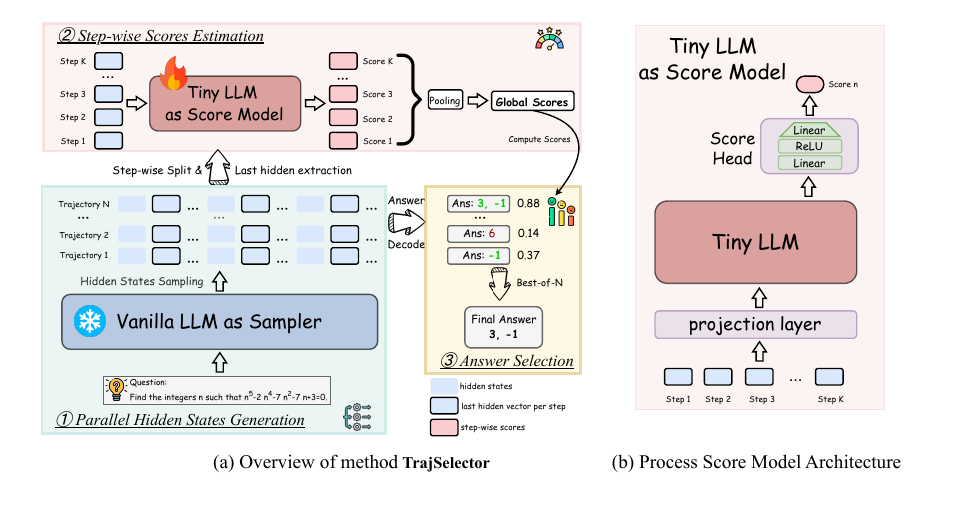}
    \captionsetup{aboveskip=4pt}
    \caption{Overview Architecture}
    \label{fig:overview}
\end{figure*}

\section{Method}
\label{sec:method}
\textbf{Key motivations}. We identify three underexplored challenges in existing external test-time-scaling (TTS) approaches: \textbf{(i)} insufficient utilization of latent cognitive processes, particularly the absence of introspective elements such as self-reflection in semantic representation (as current methods primarily operate in lexical spaces); \textbf{(ii)} high computational costs associated with large scoring models—typically a base LLM augmented with a scoring head—and the detrimental impact of auxiliary loss functions on the core causal reasoning ability of the underlying model; \textbf{(iii)} label noise introduced by automated, step-level annotation procedures.

Our primary objective is to develop a compact yet effective response selection method for Best-of-$N$ paradigm. The proposed approach trains a lightweight model (e.g., Qwen3-0.6B-Base) to score step-level latent reasoning, while keeping the primary reasoning model (e.g., Qwen3-8B) frozen. The step-level scores are then aggregated to derive a final response-level score. To improve training stability and reduce the impact of label noise, we introduce a customized classification loss coupled with pseudo-label crafting.


\subsection{Overview}
\label{sec:method:overview}

In this section, we propose a unified framework comprising a sampler model $\mathcal{M}_{\phi}$ for response generation, a lightweight  process score model ${f_\theta}$ for evaluation. This design exploits the sampler’s intrinsic latent representations—specifically, its hidden states—to inform the scoring process. Another advantage of this approach lies in its efficiency: Best-of-$N$ is achieved with minimal additional parameters, as the large sampler model remains frozen during training.

As illustrated in Figure~\ref{fig:overview}(a), the framework \emph{\textbf{\mname{}}} operates in a three-stage pipeline. First, the sampler model generates multiple candidate responses in parallel for a given query, while the last hidden states are extracted as latent representations. Second, the reasoning trajectory is segmented into discrete steps, the hidden states of which are then passed to the compact process score model, which outputs a scalar estimate of reasoning quality. These step-level scores are then aggregated to yield a global score representing the overall quality of the response. Finally, the candidate with the highest global score is selected as the optimal output.

\subsection{Process Score Model}
\label{sec:method:psm}

The process score model assigns a score between 0 and 1 to each reasoning step within a reasoning trajectory, as shown in Figure~\ref{fig:overview}(b). Architecturally, it comprises a tiny LLM and a score head. The LLM component is a 0.6B-parameter language model (Qwen3-0.6B-Base). As observed by \citet{guo2025miningintrinsicrewardsllm, DeepConf}, LLMs encode capabilities such as self-rewarding and self-reflection inherent in their hidden states. Accordingly, rather than relying on the generated tokens after a classification head for each reasoning step, the process score model takes the last hidden states as input. The score head is a shallow neural network consisting of two linear layers, producing a three-class classification output: \emph{wrong}, \emph{neutral}, and \emph{right}. The \emph{neutral} class serves as a noise-absorbing buffer, which will be elaborated upon in the following sections. The model’s final score for a reasoning step is the predicted probability assigned to the \emph{right} class by the scoring head.

From a runtime perspective, the process involves three components: (i) segmenting the reasoning trajectory into discrete steps; (ii) scoring each step using the process score model; and (iii) aggregating the individual scores to compute an overall response score.

\textbf{Segmentation of reasoning steps}. The reasoning trajectory, enclosed between the \texttt{<think>} and \texttt{</think>} tags in each response, is extracted for step segmentation. We divide this trajectory into discrete steps using the delimiter \texttt{’\textbackslash n\textbackslash n’}, thereby avoiding the need to introduce additional step-specific tokens or fine-tune the LLM to produce step-formatted outputs. After segmentation, the final token of each step is identified, and its last hidden state from the sampler model is used for scoring.

\textbf{Step-wise scoring}. For each reasoning step, the hidden states of its step token serves as the self-reflective signal of that step. These representations are concatenated into a sequence and input into the process score model. The model analyzes the sequence and produces a score between 0 and 1 for each step via its score head.

\textbf{Score aggregation}. Step-wise scores are combined through a pooling operation to compute a global score representing the overall quality of the reasoning trajectory. Specifically, we employ an arithmetic mean as the pooling function.

Following aggregation, each response is assigned a global trajectory score, with higher scores reflecting higher-quality reasoning. The response with the highest global score is selected as the final Best-of-$N$ output.

\subsection{Training}
\label{sec:method:training}

A central challenge in training the process score model lies in the scarcity of high-quality labeled data for intermediate reasoning steps. To address this, we adopt the strategy introduced in FreePRM \citep{FreePRM}, casting the training process as a standard classification task that employs only the final outcome label as weak supervision. To further alleviate label noise, we incorporate an auxiliary mechanism designed to absorb uncertainty in the supervision signal. This approach obviates the need for labor-intensive, manually annotated intermediate steps and supports a data-driven paradigm in which the model learns to evaluate process quality autonomously.

Given a reasoning trajectory $\tau = (s_1, s_2, \cdots, s_T)$ consisting of $T$ steps, and a ground truth label $y \in \{0, 1 \}$ indicating whether the final answer is correct, we create a pseudo-label $\tilde{y}_t$ for each step $s_t \in \tau$ as defined in Equation~\ref{eq:pseudo-label}.

\begin{equation}\label{eq:pseudo-label}
    \tilde{y}_t = y, ~~ t = 1, 2, \ldots, T
\end{equation}

However, this pseudo-labeling strategy introduces step-level noise, as not all steps within a trajectory that leads to a correct final answer are necessarily high-quality. To mitigate this, we extend the binary classification task by introducing a third class as a buffer \citep{FreePRM}. Accordingly, for each reasoning step $s_t$, the process score model predicts a probability distribution over three classes: right ($\textcolor{darkgreen}{p^r_t}$), wrong ($\textcolor{red}{p^w_t}$), and buffer ($\textcolor{blue}{p^b_t}$), subject to the constraint defined in Equation~\ref{eq:classify-constraint}.
\begin{equation}\label{eq:classify-constraint}
\begin{aligned}
& (\textcolor{darkgreen}{p^r_t}, \textcolor{red}{p^w_t}, \textcolor{blue}{p^b_t}) = f_\theta(s_t) \\
& ~~ s.t. ~ \textcolor{darkgreen}{p^r_t} + \textcolor{red}{p^w_t} + \textcolor{blue}{p^b_t} = 1
\end{aligned}
\end{equation}
where the distribution is obtained with a softmax transformation on the output embedding of the scoring head.

Accordingly, for a pseudo-label $\tilde{y}_t$, the training objective is formulated to encourage the behavior specified in Equation~\ref{eq:classify-objective}:

\begin{equation}
\label{eq:classify-objective}
    \begin{cases}
        \textcolor{darkgreen}{p^r_t} + \textcolor{blue}{p^b_t} = 1, & \text{if } \tilde{y}_t = 1, \\
        \textcolor{red}{p^w_t} + \textcolor{blue}{p^b_t} = 1, & \text{if } \tilde{y}_t = 0.
    \end{cases}
 \end{equation}
    
This formulation allows the model to route ambiguous or noisy reasoning steps through the buffer class, reducing overfitting to potentially incorrect labels. Based on this objective, the final training loss is defined in Equation~\ref{eq:final-loss}, where $T$ denotes the number of reasoning steps in the trajectory $\tau$:

\begin{equation}
\label{eq:final-loss}
    \small
    \mathcal{L}(\theta | \tau) = -\frac{1}{T} \sum_{t=1}^T [\tilde{y}_t \log (\textcolor{darkgreen}{p^r_t} + \textcolor{blue}{p^b_t}) + (1 - \tilde{y}_t) \log (\textcolor{red}{p^w_t} + \textcolor{blue}{p^b_t})]
\end{equation}

During the training procedure, the sampler LLM $\mathcal{M}_\phi$ remains frozen.
\section{Experiment}
\label{sec:experiment}


\begin{table*}[!htbp]
    \centering
    \small
    \begin{tabular}{lccccccc}
    \hline
    
    \rule[-0.9ex]{0pt}{3.4ex} \textbf{Method} & \multicolumn{1}{c}{\textbf{AMC-23}} & \multicolumn{1}{c}{\textbf{AIME-24}} & \multicolumn{1}{c}{\textbf{AIME-25}} & \multicolumn{1}{c}{\textbf{BeyondAIME}} & \multicolumn{1}{c}{\textbf{HMMT-25}} & \multicolumn{1}{c}{\textbf{BRUMO-25}} & \multicolumn{1}{c}{\textbf{Avg}} \\
    & \multicolumn{1}{c}{(/40)} & \multicolumn{1}{c}{(/30)} & \multicolumn{1}{c}{(/30)} & \multicolumn{1}{c}{(/100)} & \multicolumn{1}{c}{(/30)} & \multicolumn{1}{c}{(/30)} & \multicolumn{1}{c}{(\%)} \\
    \hline

    \rowcolor{navyblue!10}\multicolumn{8}{c}{{\textit{\textbf{Best-of-32}}}} \\
    \passk{Pass@32 (Oracle)} & \passk{38} & \passk{24} & \passk{22} & \passk{46} & \passk{15} & \passk{23} & \passk{71.83} \\
    Random Selection & 34 & 17 & 12 & 17 & 8 & 16 & 46.44 \\
    Majority Voting & 36 & 20 & 17 & 25 & 8 & \textbf{18} & 54.17 \\
    ReasonFlux-PRM-7B & 35 & 19 & 16 & 21 & 8 & 16 & 50.86 \\
    Qwen2.5-Math-PRM-7B & 35 & \textbf{21} & 16 & 23 & 8 & 16 & 52.31 \\
    Qwen2.5-Math-7B-PRM800K & 36 & 19 & 15 & 20 & 7 & 15 & 49.44 \\
    ReasonEval-7B & 35 & 20 & 15 & 20 & 7 & \textbf{18} & 51.25 \\
    Math-Shepherd & 34 & 11 & 16 & 23 & 7 & 14 & 46.67 \\
    AceMath-7B-RM & 35 & 19 & 13 & 21 & 6 & 16 & 48.08 \\
    EurusPRM & 36 & 19 & 17 & 28 & 10 & 17 & 54.47 \\
    \textbf{\emph{\mname{}}} (ours) & \textbf{38} & \textbf{21} & \textbf{18} & \textbf{31} & \textbf{11} & \textbf{18} & \textbf{58.78} \\
    \hline

    \rowcolor{navyblue!10}\multicolumn{8}{c}{{\textit{\textbf{Best-of-16}}}} \\
    \passk{Pass@16 (Oracle)} & \passk{38} & \passk{24} & \passk{22} & \passk{41} & \passk{15} & \passk{21} & \passk{62.67} \\
    Random Selection & 33 & 19 & 13 & 20 & 6 & 13 & 45.42 \\
    Majority Voting & 36 & \textbf{20} & \textbf{16} & 25 & 9 & 16 & 53.06 \\
    ReasonFlux-PRM-7B & 35 & 17 & 13 & 22 & 9 & \textbf{19} & 50.47 \\
    Qwen2.5-Math-PRM-7B & 36 & 17 & 13 & 23 & 6 & 18 & 48.83 \\
    Qwen2.5-Math-7B-PRM800K & 35 & 17 & 14 & 19 & 9 & 18 & 49.97 \\
    ReasonEval-7B & 36 & \textbf{20} & 13 & 22 & 7 & 16 & 49.97 \\
    Math-Shepherd & 34 & 18 & 12 & 22 & 5 & 16 & 46.17 \\
    AceMath-7B-RM & 35 & 18 & 13 & 20 & 7 & 16 & 47.91 \\
    EurusPRM & 36 & 19 & 13 & 26 & \textbf{10} & 18 & 52.67 \\
    \textbf{\emph{\mname{}}} (ours) & \textbf{37} & \textbf{20} & \textbf{16} & \textbf{29} & \textbf{10} & 18 & \textbf{55.81} \\
    \hline
    
    \end{tabular}
    \caption{Experimental Results of Best-of-$32$ \& Best-of-$16$}
    \label{tab:main-results-1}
\end{table*}

\subsection{Experiment Settings}
\label{sec:experiment:settings}

\textbf{Training Dataset} We construct the training corpus from two public datasets: OpenR1-Math-220k \citep{openr1} and DeepMath-103K \citep{deepmath}. Each example in the public datasets contains a question and a ground-truth answer. We employ Qwen3-8B \citep{qwen3technicalreport} to perform reasoning and generate the corresponding thinking trajectory and response for each question. Each response is automatically labeled by comparing its final answer with the ground truth via \texttt{Math-Verify} \citep{math-verify}. Given the substantial imbalance where correct samples outnumber incorrect ones, we retain all incorrect samples as negatives and downsample correct samples to a 1:1 ratio for positives. The resulting dataset is then shuffled to form the final training set which contains 133K examples. Statistical information can be found in Appendix~\ref{appendix:training-data}.

\textbf{Network Architecture.} We employ Qwen3-8B as the frozen sampler model, initialize the weights of the process score model using Qwen3-0.6B-Base, implement hidden states mapping between the two LLMs through a linear layer, and construct the score head with two linear layers and a ReLU activation function \citep{ReLU}. Figure~\ref{fig:network-code} illustrates the network architecture pseudo-code.

\definecolor{codegreen}{rgb}{0,0.5,0}
\definecolor{codeblue}{rgb}{0.25,0.5,0.5}
\definecolor{codegray}{rgb}{0.6,0.6,0.6}
\lstset{
  backgroundcolor=\color{white},
  basicstyle=\fontsize{8pt}{8pt}\fontfamily{lmtt}\selectfont,
  columns=fullflexible,
  breaklines=true,
  captionpos=b,
  commentstyle=\fontsize{8.5pt}{9.5pt}\color{codegray},
  keywordstyle=\fontsize{8.5pt}{9.5pt}\color{codegreen},
  stringstyle=\fontsize{8.5pt}{9.5pt}\color{codeblue},
  frame=tb,
  otherkeywords = {self},
}
\begin{figure}[!t]
\begin{lstlisting}[language=python]
class TrajSelector(PreTrainedModel):
  # this is the annotation
  def __init__(self, config):
    ...
    self.policy_model = load(model_name="Qwen3-8B")
    self.prm = load(model_name="Qwen3-0.6B-Base")
    self.projection = nn.Linear(
      policy_hidden_states_dim, 
      prm_hidden_states_dim
    )
    self.score_head = nn.Sequential(
      nn.Linear(prm_hidden_states_dim, d),
      nn.ReLU(),
      nn.Linear(d, num_labels),
    )
    ...
\end{lstlisting}
\vspace{-1em}
\caption{PyTorch style code of \mname.}
\label{fig:network-code}
\end{figure}

\textbf{Training Strategy.} We employ DeepSpeed \citep{ZeRO} and HuggingFace transformers \citep{HuggingFaceTransformers} to train the \textbf{\emph{\mname{}}}. The hyper-parameters are shown in Appendix~\ref{appendix:training-params}.

\textbf{Benchmarks.} We conduct Best-of-$N$ experiments on the following benchmarks: AMC23, AIME24, AIME25, HMMT25, BRUMO25 \citep{MathArena} and BeyondAIME \citep{Seed1.5-Thinking}. We prompt the sampler model to generate $N$ candidate responses in parallel for a given question, then employ different methods to select one response from the candidates as the final answer, and subsequently evaluate the correctness of this final answer to compute the accuracy metric. The user instruction used for generating candidate responses is shown in Appendix~\ref{appendix:eval-prompt}.

\textbf{Baselines.} We compare our method with the following baseline methods: 
(1) \textbf{Random Selection}: Select a response randomly from the $N$ candidates.
(2) \textbf{Majority Voting} \citep{Self-Consistency}: Select the answer that appears most frequently as the final answer.
(3) \textbf{Process Reward Model}: Select the response with the highest score computed by process reward model from the $N$ candidates. We selected the following strong process reward models as baselines: Qwen2.5-Math-PRM-7B \citep{Qwen2.5-Math-PRM-7B}, Qwen2.5-Math-7B-PRM800K \citep{ProcessBench}, ReasonFlux-PRM-7B \citep{ReasonFlux-PRM}, ReasonEval-7B \citep{ReasonEval}, Math-Shepherd \citep{Math-Shepherd}, AceMath-7B-RM \citep{AceMath} and EurusPRM \citep{EurusPRM,FreePRM}.
(4) \textbf{Pass@N}: A test is considered passed if at least one of the $N$ candidate responses is correct. The results of this method represent the theoretical \underline{upper bound} for Best-of-$N$ selection.

\subsection{Experiment Results}
\label{sec:experiment:results}

The main experimental results are presented in Table~\ref{tab:main-results-1} and Table~\ref{tab:main-results-2}.


\begin{table*}[htbp]
    \centering
    \small
    \begin{tabular}{lccccccc}
    \hline
    
    \rule[-0.9ex]{0pt}{3.4ex} \textbf{Method} & \multicolumn{1}{c}{\textbf{AMC-23}} & \multicolumn{1}{c}{\textbf{AIME-24}} & \multicolumn{1}{c}{\textbf{AIME-25}} & \multicolumn{1}{c}{\textbf{BeyondAIME}} & \multicolumn{1}{c}{\textbf{HMMT-25}} & \multicolumn{1}{c}{\textbf{BRUMO-25}} & \multicolumn{1}{c}{\textbf{Avg}} \\
    & \multicolumn{1}{c}{(/40)} & \multicolumn{1}{c}{(/30)} & \multicolumn{1}{c}{(/30)} & \multicolumn{1}{c}{(/100)} & \multicolumn{1}{c}{(/30)} & \multicolumn{1}{c}{(/30)} & \multicolumn{1}{c}{(\%)} \\
    \hline

    \rowcolor{navyblue!10}\multicolumn{8}{c}{{\textit{\textbf{Best-of-1}}}} \\
    Pass@1 & 34 & 16 & 11 & 21 & 8 & 17 & 46.56 \\
    \hline
    
    \rowcolor{navyblue!10}\multicolumn{8}{c}{{\textit{\textbf{Best-of-5}}}} \\
    \passk{Pass@5 (Oracle)} & \passk{37} & \passk{21} & \passk{14} & \passk{31} & \passk{11} & \passk{17} & \passk{55.58} \\
    Random Selection & 33 & 17 & 9 & 19 & 8 & 14 & 43.58 \\
    Majority Voting & 36 & 15 & \textbf{13} & 23 & 7 & \textbf{17} & 47.72 \\
    ReasonFlux-PRM-7B & 36 & 17 & 12 & 20 & 5 & 16 & 46.11 \\
    Qwen2.5-Math-PRM-7B & 36 & 17 & 12 & 20 & 5 & 16 & 46.11 \\
    Qwen2.5-Math-7B-PRM800K & 36 & 17 & \textbf{13} & 20 & 5 & 16 & 46.67 \\
    ReasonEval-7B & 34 & 18 & 12 & 24 & \textbf{9} & 16 & 48.72 \\
    Math-Shepherd & 36 & 16 & 11 & 20 & 5 & 15 & 44.44 \\
    AceMath-7B-RM & 36 & 17 & \textbf{13} & 19 & 4 & 15 & 45.38 \\
    EurusPRM & 36 & 18 & \textbf{13} & 21 & 7 & 16 & 48.50 \\
    \textbf{\emph{\mname{}}} (ours) & \textbf{37} & \textbf{19} & \textbf{13} & \textbf{26} & \textbf{9} & \textbf{17} & \textbf{51.97} \\
    \hline

    \rowcolor{navyblue!10}\multicolumn{8}{c}{{\textit{\textbf{Best-of-10}}}} \\
    \passk{Pass@10 (Oracle)} & \passk{37} & \passk{21} & \passk{19} & \passk{38} & \passk{12} & \passk{21} & \passk{62.31} \\
    Random Selection & 35 & 16 & 13 & 21 & 4 & 18 & 46.42 \\
    Majority Voting & 36 & \textbf{18} & 12 & 23 & 4 & 15 & 46.06 \\
    ReasonFlux-PRM-7B & 35 & \textbf{18} & 12 & 23 & 5 & \textbf{19} & 48.42 \\
    Qwen2.5-Math-PRM-7B & 35 & \textbf{18} & 13 & 23 & 5 & 18 & 48.42 \\
    Qwen2.5-Math-7B-PRM800K & \textbf{37} & 17 & 13 & 22 & 6 & 18 & 49.08 \\
    ReasonEval-7B & 36 & 16 & \textbf{15} & 19 & 6 & 15 & 47.06 \\
    Math-Shepherd & 35 & 17 & 11 & 21 & 5 & 17 & 45.86 \\
    AceMath-7B-RM & 34 & 15 & 12 & 16 & 4 & 13 & 41.28 \\
    EurusPRM & 36 & 17 & 13 & 19 & 4 & 18 & 47.06 \\
    \textbf{\emph{\mname{}}} (ours) & \textbf{37} & \textbf{18} & \textbf{15} & \textbf{27} & \textbf{9} & 18 & \textbf{53.25} \\
    \hline

    \end{tabular}
    \caption{Experimental Results of Best-of-$1$, Best-of-$5$ and Best-of-$10$}
    \label{tab:main-results-2}
\end{table*}

\textbf{Effective.} As observed from the experimental results, our method \mname{} demonstrates superior performance and stronger robustness compared to other baselines across multiple well-known benchmarks, and it can exhibit consistent performance improvements under various Best-of-$N$ settings. Specifically, in the Best-of-$32$ settings, the average accuracy of our method is 4.61\% higher than that of Majority Voting, and 4.31\% to 12.21\% higher than that of other process reward model-based methods. Similar performance improvements have also been validated by experimental results under other Best-of-$N$ settings.

\textbf{Efficient.} Beyond performance gains, our method's primary advantage is enabling process scoring with a mere 0.6B model, in contrast to baselines requiring independent 7B-scale PRMs. This substantially reduces deployment and inference costs.

\subsection{Ablations \& Analysis}

\begin{figure}[!t]
  \centering
  \includegraphics[width=0.48\textwidth, trim=0 9bp 0 0, clip]{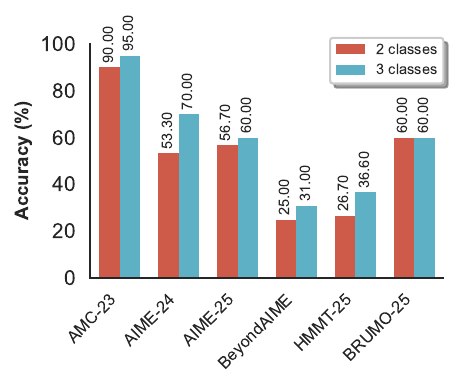}
  \captionsetup{skip=5pt}
  \caption{Ablation Study on Loss Function}
  \label{fig:ablation-loss-func}
\end{figure}

\paragraph{Ablation Study on Loss Function.} To absorb the noise information in pseudo-labels, our method modifies the standard BCELoss \citep{CrossEntropyLoss} function to design the score head as a three-class classification network. To validate the effectiveness of this design, we replace the score head with a binary classification network and employ the standard BCELoss function for training. The experimental results presented in Figure~\ref{fig:ablation-loss-func} demonstrate that the additional incorporation of a buffer probability into the design of the loss function can effectively mitigate noise in the training data and achieve enhanced performance.

\paragraph{Ablation Study on Sampler Model.} To demonstrate the generalizability of our method across different model sizes, in addition to using Qwen3-8B as the frozen sampler model, we tested the performance of Qwen3-4B and Qwen3-14B models. Appendix~\ref{appendix:ablation_policy_model} presents detailed experimental results, and these experimental findings are consistent with our conclusions.

\begin{figure}[!t]
  \centering
  \includegraphics[width=0.48\textwidth, trim=0 9bp 0 0, clip]{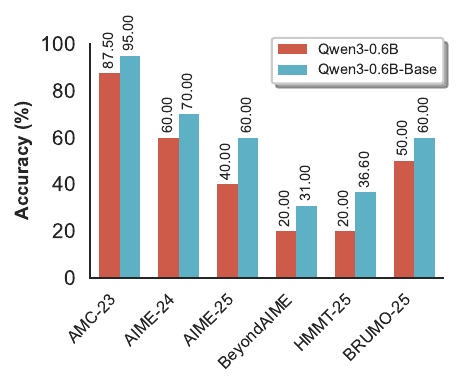}
  \captionsetup{skip=5pt}
  \caption{Ablation Study on Score Model}
  \label{fig:ablation-score-model}
\end{figure}

\paragraph{Ablation Study on Score Model.} Our method employs Qwen3-0.6B-Base as a tiny LLM to initialize the score model, which has not undergone human alignment training and thus possesses generalizability across a broader range of downstream tasks \citep{AlignmentTax}. Here, we attempt to replace it with the Qwen3-0.6B model to demonstrate its effectiveness. The results presented in Figure~\ref{fig:ablation-score-model} demonstrate that models not trained and aligned via RLHF \citep{RLHF} exhibit superior performance.

\paragraph{Larger $N$ in Best-of-$N$.} Although the range of $N$ value selection in our main experimental results has already covered common parallel sampling quantities, we attempted to increase the $N$ value to a larger extent in the Best-of-$N$ experiments. The experimental results in Appendix~\ref{appendix:larger-N} demonstrate that the effectiveness of our proposed method remains valid as the value of N increases.

\paragraph{Offline Data Selection.} To demonstrate that the trained process score model can effectively distinguish between high-quality and low-quality thinking trajectory processes, we attempted to apply the process score model to the scenario of offline data selection. More details can be found in Appendix~\ref{appendix:offline-data-selection}.
\section{Discussion}
\label{discussion}


\paragraph{Advantages over Other PRMs} Our process score model has a 0.6B parameter overhead, substantially smaller than that of mainstream 7B PRMs. Unlike conventional PRMs—where token-level information flow with the policy model requires tokenizer conversion—our approach uses hidden states for information exchange, effectively preserving adequate self-reflective signals from the sampler model.


\paragraph{Why Does the Method Work?} We attribute \mname{}’s effectiveness to three key factors: (1) Correctness checking is simpler than problem-solving and requires no model as large as the sampler; (2) It fully exploits the sampler model’s latent representations as self-reflective signals, enabling the score model to reuse part of its capabilities (detailed earlier); (3) Recent studies \citep{MetaStone-S1,KnowDontKnowUncertainty,InstrinsicRewards,DeepConf,AggLM} confirm that such self-reflective signals reflect answer correctness, laying the foundation for our method.


\paragraph{Why not use BERT?} While the heavily pretrained encoder-only BERT \citep{BERT} is well-suited for numerical regression, our experiments showed that step counts in LLM-generated thinking trajectories often exceed its 512-token context limit—leading us to abandon this approach. Instead, the growing maturity of LLM-based regression research \cite{Qwen2.5-Math-PRM-7B, Qwen3Embedding, improve-embedding} inspired us to adopt a tiny LLM as the score head.

\section{Conclusion}
\label{sec:conclusion}

In this paper, we introduced \emph{\textbf{\mname{}}}, an efficient and effective framework for Best-of-$N$ selection in large reasoning models. By exploiting the sampler LLM's intrinsic latent representations and integrating a lightweight process verifier, our approach enables step-wise scoring and trajectory aggregation without relying on costly step-level annotations or heavy standalone PRMs. The data-driven end-to-end training recipe, incorporating a noise-absorbing three-class loss, ensures robust learning and mitigates label noise. \emph{\mname{}} advances external TTS by highlighting the potential of latent representations for accessible inference. Future work could explore hybrid integration with internal TTS, extensions to non-math domains, and adaptive $N$ based on query difficulty.

\section*{Limitations}
\label{sec:limitations}

Our research primarily focuses on the Best-of-$N$ setting within external Test-time Scaling, which represents the most common scenario. However, the potential for further integration with Monte Carlo search to unlock the full effectiveness of process scoring remains underexplored. To facilitate rigorous answer verification, we concentrate on training and evaluation in the mathematical domain; extending the benefits of this external Test-time Scaling to open-ended question answering domains warrants additional research. Furthermore, this paper only makes preliminary attempts at leveraging \mname{} for selecting high-quality reasoning data, with more refined recipes awaiting future investigations.

\section*{Ethics Statement}
\label{sec:ethics-statement}

The datasets (OpenR1-Math-220K \citep{openr1} and DeepMath-103K \citep{deepmath}) and models (Qwen series \citep{qwen3technicalreport}) employed in this study are all open-source, thereby incurring no risks associated with licensing. Furthermore, as our research is centered on the mathematical domain, it does not entail risks pertaining to human ethics and values.

\bibliography{main}

\appendix

\section{Evaluation Prompt Template}
\label{appendix:eval-prompt}

The prompt template used is sourced from LightEval \citep{LightEval}.

\begin{tcolorbox}[breakable, colback=gray!5, colframe=gray!80!black, title=\textbf{Evaluation Prompt Template}, boxrule=1pt, arc=2mm, left=1mm, right=1mm, top=1mm, bottom=1mm]
    \small
    Solve the following math problem efficiently and clearly. Please reason step by step, and put your final answer within \verb|\|boxed\{\}.
\end{tcolorbox}

\section{Training Dataset Statistics}
\label{appendix:training-data}

The size of the training dataset is shown in the Table~\ref{tab:training-data-statistics}. As can be seen, the ratio of positive to negative samples we selected is maintained at 1:1.

\begin{table}[h]
    \centering
    \small
    \begin{tabular}{l|c|c|c}
    \hline
    \textbf{Data Source} & \textbf{Positive} & \textbf{Negative} & \textbf{Total} \\
    \hline
    DeepMath-103K & 11,592 & 11,592 & 23,184 \\
    OpenR1-Math-220k & 55,258 & 55,258 & 110,516 \\
    \hline
    \textbf{Overall} & \textbf{66,850} & \textbf{66,850} & \textbf{133,700} \\
    \hline
    \end{tabular}
    \caption{Training Dataset Size}
    \label{tab:training-data-statistics}
\end{table}

\section{Sampling Parameters}
\label{appendix:sampling-params}

Table~\ref{tab:sampling-params} presents the sampling parameters used during the policy model's inference time in the dataset construction process and benchmark evaluation. These parameter values are all derived from the official Best Practices recommended by Qwen3-8B \citep{qwen3technicalreport}. The inference service for the LLM is deployed using vLLM \citep{vLLM} as the foundation infrastructure.

\begin{table}[h]
    \centering
    \small
    \begin{tabular}{l|c}
    \hline
    \textbf{Parameter} & \textbf{Value} \\
    \hline
    Enable Thinking & True \\
    Temperature & 0.6 \\
    Top-p & 0.95 \\
    Max tokens & 10,000 \\
    Top-k & 20 \\
    Min-p & 0 \\
    \hline
    \end{tabular}
    \caption{Sampling Parameters}
    \label{tab:sampling-params}
\end{table}

\section{Training Parameters}
\label{appendix:training-params}

Table~\ref{tab:training-params} presents the training hyper-parameters used in the training process.

\begin{table}[h]
    \centering
    \small
    \begin{tabular}{l|c}
        \hline
        \textbf{Parameter} & \textbf{Value} \\
        \hline
        Samples & 133K \\
        Trainable Part & Full model \\
        Learning Rate & $1 \times 10^{-4}$ \\
        Epoch & 3 \\
        Optimizer & AdamW \\
        DeepSpeed & Zero2 \\
        Weight Decay & 0.1 \\
        Max Seq.Length & 10,000 tokens \\
        Per-device Batch Size & 2 \\
        Gradient Accumulation & 4 \\
        Max tokens & 10,000 \\
        Mixed Precision & bfloat16 \\
        GPU Nums & $8 \times$ NVIDIA H100 \\
        \hline
        \end{tabular}
        \caption{Training Parameters}
        \label{tab:training-params}
\end{table}

\section{Offline Data Selection}
\label{appendix:offline-data-selection}

We used OpenThoughts-114K \citep{OpenThoughts} as the source dataset, from which 1K samples were selected as the supervised fine-tuning (SFT) dataset via different data selection strategies. Subsequently, we performed SFT on the Qwen2.5-14B-Instruct \citep{qwen2.5} model; the performance of the fine-tuned model was evaluated to assess the effectiveness of the various data selection strategies. Additionally, we employed two datasets for comparison: s1K \citep{s1} (a carefully human-curated dataset) and a randomly selected dataset. All training strategies and parameter selections follow the settings in the s1 \citep{s1}, and the evaluation methods strictly adhere to the default strategy of the open-r1 \citep{openr1} evaluation suite. The rationale for this approach is that these experimental protocols have been explored in recent related studies \citep{ReasonFlux-PRM, LS-Mixture}. The experimental results are presented in Table~\ref{tab:offline-data-selection-eval}.

\begin{table}[!htbp]
    \centering
    \small
    \setlength{\tabcolsep}{1pt}
    \begin{tabular}{@{}ccccc@{}} 
    \toprule
    \textbf{Dataset} & \textbf{MATH500} & \textbf{AIME24} & \textbf{AIME25} & \textbf{GPQA} \\
    \midrule
    random & 71.6 & 16.7 & 20.0 & 34.8 \\
    s1K & 78.8 & 40.0 & 33.3 & 41.4 \\
    Math-Shepherd-PRM-7B & 67.8 & 13.3 & 6.7 & 33.3 \\
    Qwen2.5-Math-PRM-7B & 73.2 & 26.7 & 20.0 & 39.4 \\
    ReasonFlux-PRM-7B & 84.8 & 40.0 & 33.3 & 47.5 \\
    \mname{} (our) & \textbf{86.4} & \textbf{43.3} & \textbf{43.3} & \textbf{53.5} \\
    \bottomrule
    \end{tabular}
    \caption{Offline Data Selection Evaluation}
    \label{tab:offline-data-selection-eval}
\end{table}

As observed from the experimental results, the training samples selected by \mname{} are more effective in improving model performance, which reflects \mname{}’s capability to assess the quality of thinking trajectories.

\section{Ablation Study on Sampler Model}
\label{appendix:ablation_policy_model}

To explore the feasibility of using models with more diverse scales as frozen sampler models, we conducted experimental evaluations using Qwen3-4B and Qwen3-14B respectively as the models for response generation. The experimental results are presented in the Table~\ref{tab:ablation-policy-model_qwen3_4b}. From the experimental results, the adoption of \mname{} still yields benefits compared to Majority Voting method.

\begin{table*}[!htbp]
    \centering
    \small
    \begin{tabular}{lccccccc}
    \hline
    
    \rule[-0.9ex]{0pt}{3.4ex} \textbf{Method} & \multicolumn{1}{c}{\textbf{AMC-23}} & \multicolumn{1}{c}{\textbf{AIME-24}} & \multicolumn{1}{c}{\textbf{AIME-25}} & \multicolumn{1}{c}{\textbf{BeyondAIME}} & \multicolumn{1}{c}{\textbf{HMMT-25}} & \multicolumn{1}{c}{\textbf{BRUMO-25}} & \multicolumn{1}{c}{\textbf{Avg}} \\
    & \multicolumn{1}{c}{(/40)} & \multicolumn{1}{c}{(/30)} & \multicolumn{1}{c}{(/30)} & \multicolumn{1}{c}{(/100)} & \multicolumn{1}{c}{(/30)} & \multicolumn{1}{c}{(/30)} & \multicolumn{1}{c}{(\%)} \\
    \hline

    \rowcolor{navyblue!10}\multicolumn{8}{c}{{\textit{\textbf{Best-of-32 + Qwen3-4B as Sampler}}}} \\
    \passk{Pass@32 (Oracle)} & \passk{39} & \passk{24} & \passk{21} & \passk{50} & \passk{17} & \passk{22} & \passk{71.25} \\
    Majority Voting & \textbf{38} & 21 & 16 & 26 & 12 & 17 & 56.83 \\
    \textbf{\emph{\mname{}}} (ours) & \textbf{38} & \textbf{22} & \textbf{17} & \textbf{30} & \textbf{12} & \textbf{18} & \textbf{59.17} \\
    \hline

    \rowcolor{navyblue!10}\multicolumn{8}{c}{{\textit{\textbf{Best-of-32 + Qwen3-14B as Sampler}}}} \\
    \passk{Pass@32 (Oracle)} & \passk{39} & \passk{24} & \passk{22} & \passk{53} & \passk{17} & \passk{24} & \passk{73.42} \\
    Majority Voting & 39 & 21 & 14 & 34 & 13 & 21 & 60.25 \\
    \textbf{\emph{\mname{}}} (ours) & \textbf{39} & \textbf{23} & \textbf{18} & \textbf{33} & \textbf{13} & \textbf{19} & \textbf{67.86} \\
    \hline


    \end{tabular}
    \caption{Experimental Results of Best-of-$32$ on Qwen3-4B and Qwen3-14B}
    \label{tab:ablation-policy-model_qwen3_4b}
\end{table*}

\section{Larger $N$ in Best-of-$N$}
\label{appendix:larger-N}

The experimental results for Best-of-$48$ and Best-of-$64$ are presented in the Table~\ref{tab:ablation-larger-N}. The evaluation setup is consistent with Section~\ref{sec:experiment:settings} in the main experiment. 

From the experimental results, \mname{} maintains stable performance and continues to yield benefits when $N$ takes larger values, demonstrating a scaling trend where its effectiveness improves as $N$ increases.

\begin{table*}[!htbp]
    \centering
    \small
    \begin{tabular}{lccccccc}
    \hline
    
    \rule[-0.9ex]{0pt}{3.4ex} \textbf{Method} & \multicolumn{1}{c}{\textbf{AMC-23}} & \multicolumn{1}{c}{\textbf{AIME-24}} & \multicolumn{1}{c}{\textbf{AIME-25}} & \multicolumn{1}{c}{\textbf{BeyondAIME}} & \multicolumn{1}{c}{\textbf{HMMT-25}} & \multicolumn{1}{c}{\textbf{BRUMO-25}} & \multicolumn{1}{c}{\textbf{Avg}} \\
    & \multicolumn{1}{c}{(/40)} & \multicolumn{1}{c}{(/30)} & \multicolumn{1}{c}{(/30)} & \multicolumn{1}{c}{(/100)} & \multicolumn{1}{c}{(/30)} & \multicolumn{1}{c}{(/30)} & \multicolumn{1}{c}{(\%)} \\
    \hline

    \rowcolor{navyblue!10}\multicolumn{8}{c}{{\textit{\textbf{Best-of-48}}}} \\
    \passk{Pass@48 (Oracle)} & \passk{38} & \passk{24} & \passk{24} & \passk{46} & \passk{16} & \passk{23} & \passk{71.83} \\
    Random Selection & 32 & 17 & 13 & 19 & 7 & 14 & 44.83 \\
    Majority Voting & 36 & 22 & 17 & 25 & 11 & 19 & 57.50 \\
    ReasonFlux-PRM-7B & 36 & 18 & 15 & 21 & 8 & 16 & 50.17 \\
    Qwen2.5-Math-PRM-7B & 35 & 19 & 15 & 23 & 7 & 16 & 50.08 \\
    Qwen2.5-Math-7B-PRM800K & 36 & 18 & 15 & 20 & 9 & 16 & 50.56 \\
    ReasonEval-7B & 35 & 21 & 14 & 20 & 8 & 16 & 50.69 \\
    Math-Shepherd & 34 & 11 & 14 & 23 & 7 & 13 & 43.00 \\
    AceMath-7B-RM & 35 & 20 & 11 & 21 & 5 & 16 & 46.97 \\
    EurusPRM & 36 & 20 & 17 & 27 & 12 & 18 & 56.72 \\
    \textbf{\emph{\mname{}}} (our) & \textbf{38} & \textbf{22} & \textbf{19} & \textbf{33} & \textbf{12} & \textbf{19} & \textbf{61.33} \\
    \hline

    \rowcolor{navyblue!10}\multicolumn{8}{c}{{\textit{\textbf{Best-of-64}}}} \\
    \passk{Pass@64 (Oracle)} & \passk{39} & \passk{25} & \passk{24} & \passk{55} & \passk{17} & \passk{23} & \passk{74.86} \\
    Random Selection & 36 & 15 & 13 & 25 & 4 & 14 & 44.72 \\
    Majority Voting & 37 & 23 & 17 & 35 & 13 & 19 & 61.25 \\
    ReasonFlux-PRM-7B & 34 & 20 & 17 & 29 & 9 & 17 & 54.00 \\
    Qwen2.5-Math-PRM-7B & 34 & 20 & 16 & 27 & 9 & 16 & 52.56 \\
    Qwen2.5-Math-7B-PRM800K & 35 & 19 & 16 & 28 & 9 & 16 & 52.58 \\
    ReasonEval-7B & 35 & 18 & 15 & 22 & 7 & 16 & 49.36 \\
    Math-Shepherd & 33 & 17 & 17 & 22 & 7 & 15 & 48.53 \\
    AceMath-7B-RM & 33 & 16 & 14 & 20 & 5 & 18 & 46.52 \\
    EurusPRM & 37 & 21 & 18 & 30 & 11 & 19 & 58.75 \\
    \textbf{\emph{\mname{}}} (our) & \textbf{39} & \textbf{23} & \textbf{19} & \textbf{37} & \textbf{12} & \textbf{20} & \textbf{63.52} \\
    \hline
    
    \end{tabular}
    \caption{Experimental Results of Best-of-$48$ \& Best-of-$64$}
    \label{tab:ablation-larger-N}
\end{table*}



\end{document}